\documentclass[runningheads]{llncs}
\usepackage{graphicx}
\usepackage{subfigure}
\usepackage{caption}
\usepackage{subcaption}
\usepackage{float}
\usepackage{xcolor, soul}
\usepackage{algorithm}
\usepackage{algorithmic}
\usepackage{mwe}
\usepackage{graphicx}
\usepackage{subcaption}
\usepackage{hyperref}
\usepackage[strings]{underscore}
\usepackage{underscore}

\begin{document}

\title{Aligning Brain Activity with Advanced Transformer Models: Exploring the Role of Punctuation in Semantic Processing}

\author{%
Zenon Lamprou \inst{1} \orcidID{0000-0003-0042-5036} \and 
Frank Pollick \inst{2} \orcidID{0000-0002-7212-4622} \and 
Yashar Moshfeghi \inst{1} \orcidID{0000-0003-4186-1088}
}%
\institute{
NeuraSearch Laboratory, University of Strathclyde, Glasgow, UK\\
\email{zenon.lamprou@strath.ac.uk}, \email{yashar.moshfeghi@strath.ac.uk}
\and
University of Glasgow, Glasgow, UK\\
\email{frank.pollick@gla.ac.uk}\\ 
}
\authorrunning{Lamprou et al.}



\maketitle

\begin{abstract}

This research examines the congruence between neural activity and advanced transformer models, emphasizing the semantic significance of punctuation in text understanding. Utilizing an innovative approach originally proposed by Toneva and Wehbe, we evaluate four advanced transformer models—RoBERTa, DistiliBERT, ALBERT, and ELECTRA—against neural activity data. Our findings indicate that RoBERTa exhibits the closest alignment with neural activity, surpassing BERT in accuracy. Furthermore, we investigate the impact of punctuation removal on model performance and neural alignment, revealing that BERT's accuracy enhances in the absence of punctuation. This study contributes to the comprehension of how neural networks represent language and the influence of punctuation on semantic processing within the human brain.

\keywords{fMRI  \and Transformers \and Explainable AI \and Punctuation Symbols}
\end{abstract}

\section{Introduction}
Continuous advancements in the area of natural language processing (NLP) have produced advanced deep learning models. How these models represent language internally is currently not very well understood because these models often tend to be large and complex. However, what is clear is that these deep learning models don't learn from any explicit rules regarding language, rather they learn generic information about language itself. Some of these models are trained on a large generic corpus and then fine-tuned in a downstream task. These models tend to perform better than models that are trained on just one specific NLP task \cite{devlin_bert_2019,peters-etal-2018-deep,howard-ruder-2018-universal}.

There has been much effort to explain the inner representations of such models. Some of these efforts have focused on designing specific NLP tasks that target specific linguistic information \cite{zhu-etal-2018-exploring,conneau-etal-2018-cram,linzen_assessing_2016}. 

Other research has tried to offer a more theoretical explanation regarding what word embeddings can represent \cite{peng_rational_2018,chen-etal-2018-recurrent,weiss_practical_2018}.
Toneva and Whebe \cite{toneva_interpreting_2019} in their effort to explain the inner representation of four NLP models proposed a novel approach that used brain recordings obtained from functional magnetic resonance (fMRI) and magnetoencephalography (MEG). They hypothesised that by aligning brain data with the extracted features of a model, the mapping from brain to model representation can be learned. This mapping  primarily can then offer insights on how these models represent language. They refer to this approach as aligning neural network representation with brain activity.

 One of the models tested by \cite{toneva_interpreting_2019} was BERT. BERT, since its introduction \cite{devlin_bert_2019} has generated a great deal of related research. Much of this work has been done to either improve its performance or to fix particular aspects of BERT or to fine-tune BERT to produce a version of BERT expert in a particular subject domain \cite{beltagy_scibert_2019,jia_rmbert_2021,hong_gilbert_2021,sanh_distilbert_2020,zhuang-etal-2021-robustly,lan_albert_2020,clark_electra_2020}. Toneva and Whebe \cite{toneva_interpreting_2019} were able to learn a mapping from the representations of BERT, ELMo \cite{peters-etal-2018-deep}, USE \cite{cer-etal-2018-universal} and T-XL \cite{dai_transformer-xl_2019} to brain data. However, this alignment was not tested against more sophisticated versions of BERT. We use the same novel approach as \cite{toneva_interpreting_2019} to examine four of these more sophisticated versions of the BERT model: RoBERTa \cite{zhuang-etal-2021-robustly}, DistiliBERT \cite{sanh_distilbert_2020}, ELECTRA \cite{clark_electra_2020} and ALBERT \cite{lan_albert_2020}.

 Moreover, because this novel approach is a proof of concept to show alignment of brain and neural networks we wanted to explore whether removing different punctuation symbols would influence results, potentially to produce a better alignment between brain and neural networks. Research investigating punctuation \cite{moore_whats_2016} has shown effects of punctuation on reading behaviour. Punctuation symbols can help the reader skim through text faster, but less is known about the semantic role of punctuation symbols and how they are processed in the brain. There is a substantial amount of research that examines how the brain processes text semantically and syntactically \cite{reddy_can_2021,caucheteux_language_2021,shain_constituent_2021}. In addition there is research \cite{acunzo_deep_2022} that examined which brain areas are responsible for different operations, e.g. a brain area that is responsible for understanding the meaning of complex words. However, even if the goals of these research are different, their approaches share a common fMRI experimental approach, which involves participants reading a set of sentences. When sentences appear to the participants in these experiments, a choice is made whether or not to include punctuation in the pre-processed text. Though there appears to be no general agreement if the punctuation should be included or not.

\noindent{\bf Contributions: }
The contributions of this research are divided into two parts that we present in the subsequent sections. Firstly, we examined the degree to which the four different, BERT-related transformer models aligned to brain activity. Secondly, by using the concept of learning a mapping from brain to neural network introduced by \cite{toneva_interpreting_2019} we try to use the neural network to obtain an hypothesis about the brain rather than using the brain to make observations about the network. In particular by using altered sequences, in terms of punctuation, we hypothesise that if we can get better alignment then punctuation might be processed differently for the human brain to comprehend the semantics of language.

\section{Related work}
Recent interdisciplinary research has investigated the potential role of neuroscience in enhancing information retrieval systems, a research area known as NeuraSearch \cite{moshfeghi2021neurasearch}. The findings from studies such as \cite{mueller2008electrophysiological,muller2015electroencephalography,gwizdka2019introduction,Martinez_Castano2022} indicate that NeuraSearch aims to improve access to and use of information by utilizing neurophysiological signals to provide implicit relevance feedback, thus enriching user interaction without the necessity for explicit inputs. For example, research like \cite{Moshfeghi_relevance_2013} incorporated users' emotional and physiological reactions to boost system interaction. These studies have combined various modalities, such as emotional indicators and physiological metrics like heart rate, to increase the accuracy of predicting user preferences through implicit signals. This suggests that neurophysiological tools contribute to the personalization of search experiences. 

Furthermore, research in \cite{moshfeghi_understanding_2016} employed fMRI to investigate the cognitive processes underlying information needs during question-and-answer sessions, mapping neural activities corresponding to different knowledge states. This collection of research underscores how neurological insights are pivotal in tailoring user experiences in information retrieval, as further supported by EEG studies like \cite{moshfeghi2019towards}, which identified EEG patterns linked to various stages of recognizing information needs.

The research conducted by \cite{michalkova_information_2024} investigated how brain signal integration can enhance query performance. The study demonstrated that employing electroencephalography (EEG) based models can significantly boost the relevance of search results. This result underscores the practical benefits of merging conventional querying methods with neurophysiological data to refine search efficiency.

NeuraSearch places a strong emphasis on assessing cognitive workload via EEG analysis, with studies underscoring its effectiveness in monitoring cognitive states in diverse scenarios \cite{Kingphai_Mental_2021}. Research efforts by \cite{kingphai_eeg_2021,kingphai_time_2023} have advanced in emotion recognition and classifying EEG data, underscoring the importance of methodological precision and the need for standardized procedures. Essential neuroimaging techniques like MEG and fMRI are fundamental to NeuraSearch's research \cite{kauppi2015towards}. Their utilization, as evidenced in \cite{lamprou_role_2022}, examined semantic understanding and showcased how neuroscience adeptly evaluates cognitive processing and decision-making.

The intersection of neuroscience and information retrieval reveals new potential for crafting adaptive, user-focused systems. By effectively analyzing users' implicit responses via neurophysiological indicators, as demonstrated in key research works \cite{allegretti_when_2015,Jacucci2019IntegratingNR}, it supports the development of more refined interfaces. This continuous integration seeks to meet intricate user information needs and adapt to the cognitive challenges involved in information retrieval.

A frequent design used in fMRI experiments is to contrast brain activity in two conditions \cite{friederici_brain_2011}. Often one condition is a specifically controlled baseline and the other condition reflects a process of interest. For example, to capture how the brain semantically processes information, the conditions are often a sentence vs a list of words \cite{friederici_brain_2011}. However, some research \cite{brennan_syntactic_2012,lerner_topographic_2011,wehbe_simultaneously_2014,huth_natural_2016,blank_domain-general_2017} has used a narrative to capture how the brain semantically processes information.
The work done by Toneva and Wehbe \cite{toneva_interpreting_2019} was done in a non-controlled environment and the condition they were trying to capture was how semantically the brain processes text sequences. They used these text sequences to learn a mapping from neural model features to brain features. However this approach didn't take into account how punctuation influences semantic processing but rather how the brain processes the text sequence as a whole.

 Previous work \cite{wehbe_simultaneously_2014} managed to align MEG brain activity with a Recurrent Neural Network (RNN) and its word embedding, and that informed sentence comprehension by looking at each word. Other research demonstrated that one could align a Long Short-Term Memory (LSTM) \cite{jain_incorporating_2018}  network with fMRI recordings to measure the amount of context in different brain regions. These results provided a proof of concept that one could meaningfully relate brain activity to machine learning models. Other approaches combining brain data and neural networks have tried to identify brain processing effort by using neural networks to learn a mapping from brain activity to neural network representations \cite{frank_erp_2015,hale_finding_2018}.

Previous research has explored the role of punctuation in natural language processing. From a natural language processing view, one of the standard text pre-processing steps is to remove punctuation symbols from the text to reduce the noise in the data. Some researchers have removed punctuation to achieve better results in spam detection \cite{etaiwi_impact_2017}, or plagiarism detection \cite{chong2010using}. Such results have provided mixed results in the sense that in some cases removing punctuation can lead to better results, but not every time. Recent research \cite{ek_how_2020} explored punctuation and differences in performance of MultiGenre Natural Language Inference (MNLI) using BERT and other RNN models. Their results showed that no model is capable of
taking into account cases where punctuation is meaningful, and that punctuation is generally insignificant semantically.

From a neuroscience point of view there is limited research that assesses the role of punctuation in semantics and its purpose in brain processing of text. 
Related to the role of punctuation in the syntax of language, some research has tried to focus on creating multidimensional-features that are specific to syntactical parts of language \cite{reddy_can_2021}. Then using these features they tried to model the syntactical representation of punctuation symbols in the text. There is variability in the literature in how punctuation is treated. For example, using the same fMRI data but with different research goals, one study preprocessed the text seen by the participants with punctuation \cite{caucheteux_language_2021} and the other without \cite{shain_constituent_2021}. Furthermore,  other research \cite{acunzo_deep_2022} tried to show that the combination of words can create more complex meaning and they tried to find which brain regions are responsible for representing the meaning of these words. In their pre-processing step they decided to remove the punctuation from text before presenting to their participants. This research further illustrates how there is no consensus on how punctuation is processed from the brain semantically and syntactically, and if it is needed to be included in a corpus to achieve better results.

 Overall, it can be seen that there is not enough research that evaluates or improves NLP models through brain recordings as proposed by Toneva and Wehbe \cite{toneva_interpreting_2019}. Though there is research  examining cognition that evaluates whether word embeddings contain relevant semantics \cite{sogaard-2016-evaluating}. Furthermore, other research \cite{fyshe-etal-2014-interpretable} tried to create new embeddings that align with brain recordings to asses if these new embeddings aligned better with behavioural measures of semantics. We believe that by using new state-of-the-art models to assess how their representations align with brain activity we can help expand this important and growing research area. We can unravel what training choices might lead to better alignment with brain data by assessing how the training choices of particular models improve or worsen this alignment.


\section{Methodology}
\subsection{Data}
The fMRI data were obtained originally as part of previously published research \cite{wehbe_simultaneously_2014}  and were made publicly available to use.\footnote{\scriptsize The data and the original code can be found at \href{https://github.com/mtoneva/brain_language_nlp}{this link}} Their experimental setup involved 8 subjects reading chapter 9 from Harry Potter and the Sorcerer\textquotesingle s stone \cite{bibtexkey} in its entirety. The data acquisition was done in 4 runs. Every word in the corpus appeared for 0.5s on the screen and a volume of brain activity represented by the BOLD signal would be captured every 2s (1 TR). Because of this timing arrangement, in each TR 4 words would be contained in each brain volume. These 4 words are not in random order, they are presented in the order they appear in the Harry Potter book. The data for every subject were preprocessed and smoothed by the original authors and the preprocessed version is the one available online. In their paper they note the use of MEG data as well but in this research we only focus using the fMRI since only these data have been made publicly available.

\subsection{Transformer models}
\label{sec:TransModel}
In our effort to find a more advanced transformer model that might produce better brain alignment we tested four different models against BERT, which we used as our baseline. Each model was selected based on its new characteristics compared to BERT. 
\begin{itemize}
    \item \textbf{RoBERTa} RoBERTa was one of the first variations of BERT that came after its initial introduction. \cite{zhuang-etal-2021-robustly} stated that training neural networks is often expensive and its performance really relies on hyper-parameter choices. By comparing different techniques on the two main tasks that BERT was pre-trained on, Masked Language Modelling (MLM) and Next Sentence Prediction (NSP) they were able to produce a better performing version of BERT. The interest in using RoBERTa to try and understand its inner representation is to understand if choosing different hyper-parameters in a natural language experiment can lead to better representations that are closer to how the brain represents contextual information.
    \item \textbf{DistiliBERT} Transformer model architectures using NLP tend to be large and slow to fine-tune and pre-train. \cite{sanh_distilbert_2020} in their pursuit to combat those problems proposed a BERT architecture that is reduced in size by 40\% and is 60\% percent faster to train compared to the original BERT model. Moreover during their experiments they managed to retain 97\% of its language understanding capabilities. We thought that this design choice makes DistiliBERT interesting to test and show that indeed its internal representations are not affected by the reduction in size. 
    \item \textbf{ALBERT} ALBERT was firstly introduced by \cite{lan_albert_2020} it tried to address the factor that modern day natural language processes are too big and we can easily reach memory limits when using a GPU or TPU. To combat that issue they proposed a two parameter reduction technique. The first technique they used is called factorised embedding parameterization. They hypothesised that by exposing the embedding space to a lower embedding space before exposing it to the hidden space of the model they can reduce the embedding space dimension without compromising the performance. The second technique that ALBERT implements is cross-layer parameter sharing. All the parameters in ALBERT are shared across all its layers. The features of the model in the original research \cite{toneva_interpreting_2019} are extracted for each layer independently. The prediction and evaluation on the brain data is done layer by layer using the extracted features of each layer. We hypothesise that if ALBERT's features yield a better accuracy when used on predicting brain data, then this would suggest that the brain has a uniform weighting mechanism across its own layers.
    \item \textbf{ELECTRA} ELECTRA was developed by \cite{clark_electra_2020}. Their purpose for this model was to redefine the Masked Language Modelling (MLM) task, which was one of the two tasks that the original BERT model was pre-trained on. Because modern natural language models generally require a large amount of computational power to train, they proposed an alternative to the MLM task where instead of using the [MASK] token to mask a token in the sentence and have the model predict a possible value for that token they replace it with a proposed solution generated from a small generation network. Then the main model is pre-trained by trying to predict if the generated token is indeed a correct one or if is not correct, similarly to a binary classification task. They hypothesised that this process results in a model with better contextual representations than BERT. This hypothesis makes this model a good candidate to assess its performance when aligned with brain data and determine if its inner representations are closer to how the brain represents contextual information.
\end{itemize}
\subsection{Experimental Procedure}

Our experimental procedure can be divided into three main operations. 
First we extracted the features from the desired models using different sequence lengths.
Then we trained a ridge regression model using the extracted features and used the ridge regression to predict the brain data.
Finally, we evaluated the predictions using searchlight classification.
In the rest of this section we present each of these operations in detail.The code used to run the procedure that follows was made publicly available \footnote{\scriptsize \href{https://github.com/NeuraSearch/Brain-Transformer-Mapping-Punctuation.git}{}}.

\subsubsection{Extracting features from models}
\label{expiramentalProcedure}
The first step of the experimental procedure was to extract features from different sequences of length S. In the original research \cite{toneva_interpreting_2019} the analysis was done using sequence lengths of 4, 5, 10, 15, 20, 25, 30, 35, 40 so we followed the same pattern. The sequences are formed from chapter 9 from Harry Potter and the Sorcerer\textquotesingle s Stone book. A dictionary was created that contains an entry for every layer in the model, plus the embedding layer if it exists. The representations of the embeddings layer were extracted first for every word in the chapter. The next step was to get the first sequence of S length and extract the representation of every layer for that sequence S times. Then the representation of every sequence of S length was extracted using a sliding window the same way as the first sequence. The first sequence was extracted S times to ensure that the number of sequences was the same as the number of words in the dataset. A separate script was used for every model we chose. In order to use the pre-trained checkpoints of the models the hugging face library was used so we could download this checkpoints. 

\subsubsection{Making predictions using the fMRI recordings and the extracted features}
The second step was to train a ridge regression  model  to learn a mapping from brain representation to neural network representation. This approach followed some previous work \cite{wehbe_simultaneously_2014,sudre_tracking_2012,wehbe_aligning_2014,nishimoto_reconstructing_2011,huth_natural_2016} that learned a linear function with a ridge penalty to learn the mapping from brain to neural network representation. Although a ridge regression is a relatively simple model we chose to use it since it previously has been demonstrated to be useful \cite{toneva_interpreting_2019}. However as a future work, new more complex models can be used with the hope of producing a better mapping. We used this ridge regression model to make predictions using the brain data and the features extracted from the previous step. The predictions were done for every layer, for every subject and for every model in a cross validation setting. We used a 4-fold cross validation setting for the predictions which followed the same procedure followed by Toneva and Wehbe \cite{toneva_interpreting_2019}. Because the fMRI data were gathered in four runs for every fold the fMRI data gathered in the corresponding run were treated as test data for that fold. First the extracted features from the models were loaded and using PCA were reduced to ten dimensions. The next step was to time align the corresponding fMRI data with the corresponding features extracted from the model so we knew the mapping between the features and brain data for the ridge regression to be trained on.  A total of 1351 images were recorded for every subject. By removing the edges of every run, to combat the edge effect, the 1351 were reduced to 1211. In order to align the features with the fMRI data we find out in which TR each word was presented to the user. Using that mapping because the sequences are the same number as the words then the sequence in the same index as a word corresponds to the same TR. After the mapping from TR to sequence is complete the t-4,t-3,t-2,t-1 at any given time t were concatenated to t, because by concatenating representation in the previous four time points can give better representation to the model and yield better accuracy at the end \cite{toneva_interpreting_2019}. After the previous steps were done we ended up with an array of 1351 values which are the sequences time-stamped. The last part was to remove the edges in every run, and that resulted in 1211 time-stamped sequences. Moving on, time stamping was done and the ridge regression model was trained. Different weights were obtained for every voxel using a set of different lambda values where  $\lambda=10^{x} where  -9 \geq x \leq 9$  for the ridge regression model. Then for every voxel the lambda  value  that produced the lowest error score was used to construct the weights of the final model. Then the model generated from this procedure was used to make predictions on the test data.

\subsubsection{Evaluating the predictions}
The final step was to evaluate predictions from the ridge regression model. To do that \cite{toneva_interpreting_2019}  used a searchlight classification algorithm to binary classify neighbours for every voxel, for every subject. The neighbourhoods have been pre-computed and were provided from the original research. We used the same approach and the same pre-computed neighbourhood for our evaluation method. We used the recorded brain data and a random chunk of 20 TRs was chosen and this corresponds to the correct predictions. From the predicted data the same chunk was chosen and then another random chunk of 20 TRs was chosen from the predicted data to be the wrong chunk. Then using the neighbourhoods the euclidean distance between the 2 chunks from the predicted chunk was calculated. If the distance from the correct chunk was less than the wrong chunk the prediction was marked as correct. This process was done 1000 times for every voxel and at the end the average accuracy for every voxel for every fold was obtained for every subject.

\subsubsection{Removing punctuation}
To be able to assess how the human brain processes punctuation symbols semantically we used the same experimental procedure described above but altering slightly the first part. We used four different scenarios to remove the punctuation characters before inputting the sequences to the models. Then we extracted the features in the same manner as described above but as an input we used the altered text sequences. After the sequences were extracted then the same approach was followed. The four scenarios we tested were: 
\begin{enumerate}
    \item Replace fixation ("+") symbol with the ["UNK"] token (Removing fixation).
        This scenario was used to assess if using the fixation symbol in the sequences as text can influence the quality of the extracted features and how brain aligned they are. The ["UNK"] token indicated in this scenario that this is an unknown token for the model. 
    \item Replace fixation symbol ("+") with the ["PAD"] token (Padding fixation).
     This scenario was used to assess if using the fixation symbol in the sequences as text can influence the quality of the extracted features and how brain aligned they are. The ["PAD"] token indicated to the model that this symbol should be ignored when extracting the sequences. 
    \item Replace [ "--","…","—"] with the ["PAD"] token (Padding all).
     When looking at the text from the Harry Potter chapter we identified in the text some special characters. We wanted to assess if these special characters can influence the performance when learning the brain-network mapping.
    \item Replace ["--","…","—",".","?"] with the ["PAD"]token (Padding everything).
     Motivated from the previous scenario we extended the list of the special characters by adding "." and "?" to the list of tokens to be ignored since they are one of the most commonly used punctuation symbols. Again we wanted to test how this alteration in sequences can affect the mapping.
\end{enumerate}

\section{Results and discussion}

\begin{figure}[hbt!] 
\centering
\subfigure[ALBERT]{%
\includegraphics[width=0.4\textwidth]{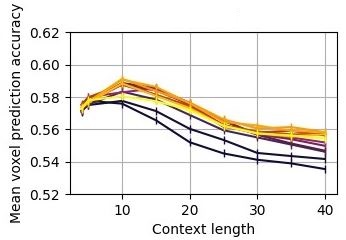}%
\label{fig:albert}%
}\hfil
\subfigure[RoBERTa]{%
\includegraphics[width=0.4\textwidth]{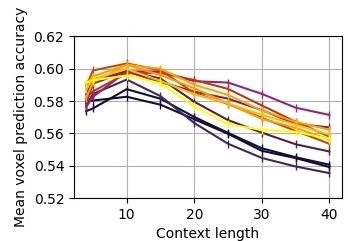}%
\label{fig:roberta}%
}

\subfigure[ELECTRA]{%
\includegraphics[width=0.4\textwidth]{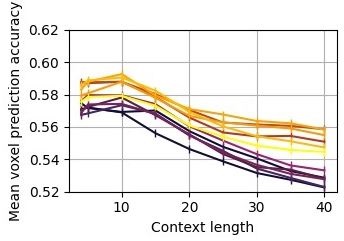}%
\label{fig:electra}%
}\hfil
\subfigure[DistiliBERT]{%
\includegraphics[width=0.4\textwidth]{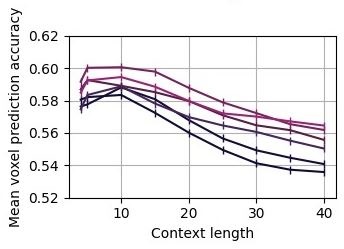}%
\label{fig:distili}%
}
\subfigure{
\includegraphics[width=0.55\textwidth]{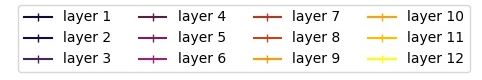}%
}

\caption{\scriptsize This figure shows the overall accuracy across all subjects for the four different models we investigated.}
    \label{fig:modelResults}
\end{figure}

\noindent{\bf Comparing the models: }
We compared the performance of BERT against the models presented in Section \ref{sec:TransModel}.
In this section we discuss the differences of these networks and what the results might show in terms of how the human brain functions. A common observation that can be seen in all the neural networks including BERT is that as the sequence length increases the accuracy of the ridge regression model decreases. All the results discussed below can be seen in Figure \ref{fig:modelResults}.

\subsubsection{BERT}
\begin{figure}[h] 
\centering
\subfigure[BERT Original Results]{%
\includegraphics[width=0.432\textwidth]{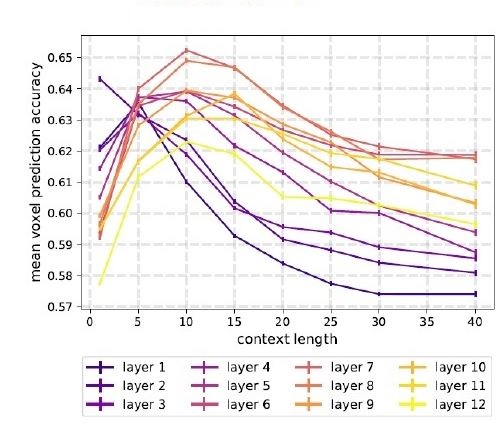}%
\label{fig:bertOriginal}%
}\hfil
\subfigure[BERT Reproduced Results]{%
\includegraphics[width=0.395\textwidth]{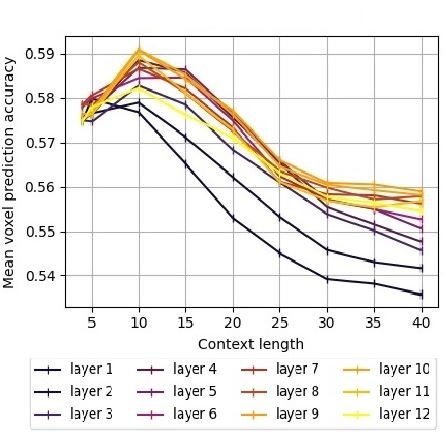}%
\label{fig:bertReproduced}%
}

\caption{\scriptsize On the left hand side are the original results reported in \cite{toneva_interpreting_2019}. On the right hand side are the reproduced results we obtained when running our code for BERT. Note that the y-axis has a different minimum and maximum value for the two panels, original and reproduced }
    \label{fig:bertResults}
\end{figure}

Following  the experimental procedure described in Section \ref{expiramentalProcedure}, and using the same data, we were not able to exactly reproduce the same original results as \cite{toneva_interpreting_2019}. Thus, our reproduced baseline, used to compare with the other models was obtained by our own code and is presented in Figure \ref{fig:bertResults} along with the original results. We believe we did not have the same results for three reasons. Firstly, when averaging across subjects and across folds for every layer the exact procedure of the averaging did not appear to be completely described in the original paper. Moreover when PCA is used the random state is not provided so we believe that this might be another reason for not replicating the exact same results. Finally our hardware is not the same as used by the original authors so this might also contribute to getting different results. However, we believe the results are accurate. By examining the results in Figure \ref{fig:bertResults} which show the original results of \cite{toneva_interpreting_2019} and our reproduced results we can see that qualitatively the overall shape of the graphs are the same but quantitatively there are some differences in the exact quantities. Similarities include that our reproduced results reaches its peak at sequence length 10, the same as the original results. Furthermore, as in the original results the accuracy decreases as the sequence length increases.

\subsubsection{ALBERT}
The first model tested was ALBERT. The interesting characteristic to assess on ALBERT was that all the layers shared the same representations. By looking closer it can be seen that there are not any major differences in the accuracy. This would suggest that the changes made on ALBERT to distinguish it from BERT have not made its representations more brain-aligned than obtained with BERT.  The fact that shared weights across the layers does not improve the alignment between brain and neural networks might suggest that the brain has a shared "weights" mechanism but this is something that needs to be investigated further.

\subsubsection{RoBERTa}
The next model tested was RoBERTa. The same trend in the graphs that is present in BERT seems to be present when using RoBERTa. An immediate observation is that RoBERTa, in almost every layer, at its peak achieves a slightly better accuracy than BERT. Taking that into consideration indicates that the decision to choose different hyper parameters when training and choosing the best performing ones leads to inner representations that are slightly better aligned with those of the human brain.

\subsubsection{DistiliBERT}
Moving on to examine the results with DistiliBERT shows that even though DistiliBERT has fewer layer, and is smaller in size than BERT the performance of its layers is better for all the layers compared to the layers of BERT. The increase is not so substantial but it indicates that the inner representation of the DistiliBERT can capture semantic information as good as BERT even though it is smaller in size. It also shows that layers of DistiliBERT are as brain aligned as the ones in BERT.

\subsubsection{ELECTRA}
Looking at the results for ELECTRA we can see that ELECTRA doesn't surpass the baseline results obtained with BERT, even though the same trend in both graphs can be seen. From this result we can argue that the training choices when training ELECTRA have not made the model as brain aligned as BERT.

\noindent{\bf Results with Removing Punctuation: }
After using the corpus without any modifications, we wanted to modify the corpus by removing punctuation symbols. In doing so, we wanted to see what this might suggest of how the brain semantically processes punctuation symbols.
\begin{figure}[h] 
\centering
\subfigure[Padding All ]{%
\includegraphics[width=0.4\textwidth]{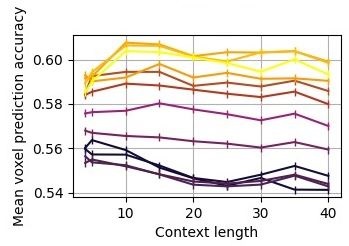}%
\label{fig:paddingAll}%
}\hfil
\subfigure[Padding Everything]{%
\includegraphics[width=0.4\textwidth]{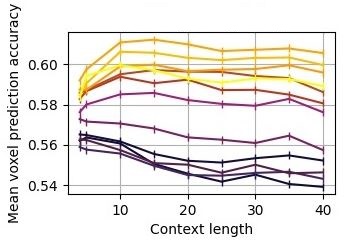}%
\label{fig:paddingEverything}%
}

\subfigure[Padding Fixations]{%
\includegraphics[width=0.4\textwidth]{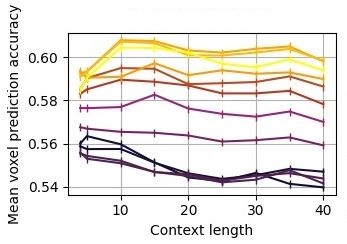}%
\label{fig:paddingFixations}%
}\hfil
\subfigure[Removing Fixations]{%
\includegraphics[width=0.4\textwidth]{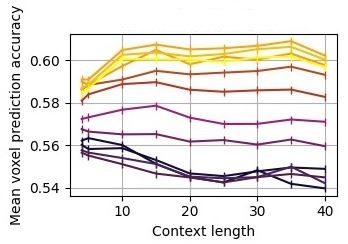}%
\label{fig:removingFixations}%
}
\subfigure{
\includegraphics[width=0.5\textwidth]{legend.JPG}%
}

\caption{\scriptsize The figures presents the results of the 4 different punctuation scenarios.}
    \label{fig:puncuationResults}
\end{figure}

Our results showed that in all four punctuation-modification scenarios the accuracy increases only on layers 7-12. The increase is not always substantial, and the biggest boost in performance is almost 1.5\%. The most interesting thing that can be seen is that removing the punctuation as padding tokens makes the model not lose as much accuracy as the length of the sequence increases. Moreover, layer 6 seems to be acting as a divisor between the starting and ending layers. This resonates with the observation of \cite{toneva_interpreting_2019} that the first 6 layers are not as brain aligned as the last 6 layers and that removing the attention mechanism from the 6 first layers might result in better representations.

Taking into account the observations mentioned above and the observation of \cite{toneva_interpreting_2019} that some layers of BERT are brain aligned, one can hypothesise that the brain might have limited use of punctuation to understand a sentence semantically. It can also be seen that as the sequence length becomes longer, which means more information is presented to the model and brain, the performance does not drop as much, revealing the limited contribution of punctuation.

\section{Conclusion}
We have investigated four different models : ELECTRA, RoBERTa, ALBERT and DistiliBERT to determine which of the models, compared to a baseline obtained with BERT, can output language representations that are more closely related to brain representations of language as indicated by the alignment between model and fMRI brain data. The most aligned models appear to be RoBERTa and DistiliBERT as they  outperform our baseline.
In addition, using the four different scenarios we wanted to test how the semantic processing of punctuation symbols might vary. By looking at the work done by Toneva and Wehbe \cite{toneva_interpreting_2019} and using their proof of concept that a mapping from the neural network features to the brain features can be learned, we tested if we can get a better mapping by removing these punctuation symbols. Indeed in all four cases the performance of the model increased. Also, the drop in accuracy when predicting brain data was not as big as originally obtained when the sequence length increased. These results support a view that the brain might have limited use for punctuation symbols to understand semantically a sentence and that as the context text length gets longer then there is less need for punctuation symbols.
%
We believe that our research can be extended in the future for more transformer models to be analysed with the goal of finding a more brain-aligned model. We also believe that the experimental procedure created by Toneva and Wehbe \cite{toneva_interpreting_2019} can be used to evaluate new models and produce models that are more brain aligned. Last, but not least we believe that the matter of how the brain processes punctuation symbols semantically should be investigated more in depth, using different models, in the quest of unravelling the capabilities of the human brain.


\bibliographystyle{splncs04}
\bibliography{biblio}
\end{document}